\documentclass{article}
\usepackage{nips07submit_e,times}

\title{Surrogate Learning - An Approach for Semi-Supervised Classification}
\usepackage{epsf}

\author{
Sriharsha Veeramachaneni\\
Thomson Reuters Research and Development\\
Eagan, MN 55123 \\
\texttt{harsha.veeramachaneni@thomsonreuters.com} \\
\And
Ravikumar Kondadadi\\
Thomson Reuters Research and Development\\
Eagan, MN 55123 \\
\texttt{ravikumar.kondadadi@thomsonreuters.com} \\
}

\begin{document}
\newcommand{\ls}[1]
   {\dimen0=\fontdimen6\the\font
    \lineskip=#1\dimen0
    \advance\lineskip.5\fontdimen5\the\font
    \advance\lineskip-\dimen0
    \lineskiplimit=.9\lineskip
    \baselineskip=\lineskip
    \advance\baselineskip\dimen0
    \normallineskip\lineskip
    \normallineskiplimit\lineskiplimit
    \normalbaselineskip\baselineskip
    \ignorespaces
   }

\newcommand{\codespacing}{1.0}
\newcommand{\linespacing}{1.5}
\newenvironment{code}
{\ls{\codespacing}
\begin{tabbing}
\rightmargin\leftmargin
\raggedright
xx\=xx\=xx\=xx\=xx\=xx\=xx\=xx\=xx\=xx\=xx\=xx\=xx\=xx \kill
}
{\end{tabbing}
}


\renewcommand{\topfraction}{.80}
\renewcommand{\bottomfraction}{.80}
\renewcommand{\textfraction}{.20}
\setcounter{topnumber}{2}
\setcounter{bottomnumber}{2}
\setcounter{totalnumber}{3}
\newcommand{\ie}{{\em i.e.}, }
\newcommand{\etal}{{\em et al. }}
\newcommand{\cell}[2]{\parbox[t]{#1}{{\raggedright #2}} \vspace*{.05in}}
\newcommand{\norm}[1]{\mbox{$\parallel #1 \parallel$}}
\newcommand{\var}[1]{\mbox{\em #1}}
\newcommand{\set}[1]{{\mathcal #1}}
\newcommand{\BM}[1]{\mbox{\boldmath $#1$}}
\newcommand{\BB}[1]{{\bf #1}}
\newcommand{\define}{\stackrel{\triangle}{=}}
\newcommand{\argmax}[1]{\raisebox{-1.5ex}{$\stackrel{\mbox{argmax}}{\scriptstyle{#1}}$ }}
\newcommand{\argmin}[1]{\raisebox{-1.5ex}{$\stackrel{\mbox{argmin}}{\scriptstyle{#1}}$ }}
\newcommand{\mymax}[1]{\raisebox{-1.0ex}{$\stackrel{\mbox{max}}{\scriptstyle{#1}}$ }}
\newcommand{\rcondition}[1]{\left. #1 \;\right|\;}
\newcommand{\lcondition}[1]{\;\left|\; #1 \right.}
\newcommand{\given}{|\,}
\newcommand{\runin}[1]{\vspace{1em} \noindent {\bf #1: }}
\newcommand{\tab}{\hspace*{0.25in}}
\newcommand{\iform}[1]{\mbox{$ #1 $}}
\newcommand{\nth}[1]{$ {#1}^{\mbox{th}}$}

\newcommand{\twobytwomatrix}[4]{\left(\begin {array}{ccc} #1 & #2 \\ #3 & #4 \end{array} \right)}
\newcommand{\twobyonevec}[2]{\left(\begin {array}{ccc} #1 \\ #2 \end{array} \right)}
\newcommand{\matrixsize}[1]{\large{#1}}

\newcommand{\BBx}{\BB{x}}
\newcommand{\BBc}{\BB{c}}
\newcommand{\BBy}{\BB{y}}
\newcommand{\BBz}{\BB{z}}
\newcommand{\BBw}{\BB{w}}
\newcommand{\BBs}{\BB{s}}
\newcommand{\BMx}{\BM{x}}
\newcommand{\BMc}{\BM{c}}
\newcommand{\BMy}{\BM{y}}
\newcommand{\BMw}{\BM{w}}
\newcommand{\BBC}{\BB{C}}
\newcommand{\BBR}{\BB{R}}
\newcommand{\BBP}{\BB{P}}
\newcommand{\mathspace}{\;}
\newcommand{\half}{\frac{1}{2}}
\newcommand{\Rvec}[1]{{\bf #1}}
\newcommand{\Ivec}[1]{\mbox{\boldmath $#1$}}
\newcommand{\Rvar}[1]{\mathsf{#1}}
\newcommand{\NewRvar}[1]{{\bf #1}}
\newcommand{\Ivar}[1]{{#1}}
\newcommand{\Matrix}[1]{\mbox{#1}}
\newcommand{\Boldmatrix}[1]{\mbox{\large{\textsf{#1}}}}
\newcommand{\matI}{\Matrix{I}_{d \times d}}
\newcommand{\mato}{\Matrix{0}_{d \times d}}

\newcommand{\Result}{\noindent {\bf Result: }}
\makeanontitle

\begin{abstract}
We consider the task of learning a classifier from the feature space $\mathcal{X}$ to the set of classes $\mathcal{Y} = \{0, 1\}$, when the features can be partitioned into class-conditionally independent feature sets $\mathcal{X}_1$ and $\mathcal{X}_2$. We show the surprising fact that the class-conditional independence can be used to represent the original learning task in terms of 1) learning a classifier from $\mathcal{X}_2$ to $\mathcal{X}_1$ and 2) learning the class-conditional distribution of the feature set $\mathcal{X}_1$. This fact can be exploited for semi-supervised learning because the former task can be accomplished purely from unlabeled samples. We present experimental evaluation of the idea in two real world applications.
\end{abstract} 

\section{Introduction}
Semi-supervised learning is said to occur when the learner exploits (a presumably large quantity of) unlabeled data to supplement a relatively small labeled sample, for accurate induction. The high cost of labeled data and the simultaneous plenitude of unlabeled data in many application domains, has led to considerable interest in semi-supervised learning in recent years.

We show a somewhat surprising consequence of class-conditional feature independence that leads to a simple semi-supervised learning algorithm. When the feature set can be partitioned into two class-conditionally independent sets, we show that the original learning problem can be reformulated in terms of the problem of learning a predictor from \emph{one} of the partitions to the other. That is, the latter partition acts as a \emph{surrogate} for the class variable. Since such a predictor can be learned from only unlabeled samples, an effective semi-supervised algorithm results.

In the next section we present the simple yet interesting result on which our semi-supervised learning algorithm (which we call \emph{surrogate learning}) is based. We present examples to clarify the intuition behind the approach and present a special case of our approach that is used in the applications section. We then examine related ideas in previous work and situate our algorithm among previous approaches to semi-supervised learning. We present empirical evaluation on two real world applications where the required assumptions of our algorithm are satisfied.

\section{Surrogate Learning}

We consider the problem of learning a classifier from the feature space $\mathcal{X}$ to the set of classes $\mathcal{Y} = \{0, 1\}$. Let the features be partitioned into $\mathcal{X} = \mathcal{X}_1 \times \mathcal{X}_2$. The random feature vector $\BBx \in \mathcal{X}$ will be represented correspondingly as $\BBx = (\BBx_1, \BBx_2)$.  Since we restrict our consideration to a two-class problem, the construction of the classifier involves the estimation of the probability $P(\BBy = 0|\BBx_1, \BBx_2)$ at every point $(\BBx_1, \BBx_2) \in \mathcal{X}$. 

We make the following assumptions on the joint probabilities of the classes and features.\\
1. $P(\BBx_1, \BBx_2 | \BBy)  = P(\BBx_1| \BBy)P(\BBx_2 | \BBy)$ for $\BBy \in \{0, 1\}$. That is, the feature sets $\BBx_1$ and $\BBx_2$ are class-conditionally independent for both classes. Note that in general our assumption is less restrictive than the \emph{Naive Bayes} assumption.\\
2. $P(\BBx_1 |\BBx_2) \neq 0$, $P(\BBx_1 |\BBy) \neq 0$ and $P(\BBx_1 |\BBy =0) \neq P(\BBx_1 |\BBy =1)$. These assumptions are to avoid \emph{divide-by-zero} problems in the algebra below. If $\BBx_1$ is a discrete valued random variable and not irrelevant for the classification task, these conditions are often satisfied.

Under these assumptions, surprisingly, we can establish that $P(\BBy=0 | \BBx_1, \BBx_2)$ can be written as a function of $P(\BBx_1|\BBx_2)$ and $P(\BBx_1|\BBy)$. First, when we consider the quantity $P(\BBy, \BBx_1| \BBx_2)$, we may derive the following.

\begin{eqnarray}
&& P(\BBy, \BBx_1| \BBx_2) = P(\BBx_1|\BBy, \BBx_2) P(\BBy| \BBx_2) \nonumber\\
&\Rightarrow& P(\BBy, \BBx_1| \BBx_2) = P(\BBx_1|\BBy) P(\BBy| \BBx_2) \;\;\;\mbox{(from the independence assumption)} \nonumber\\
&\Rightarrow& P(\BBy | \BBx_1, \BBx_2) P(\BBx_1| \BBx_2) = P(\BBx_1|\BBy) P(\BBy| \BBx_2) \nonumber\\
&\Rightarrow& \frac{P(\BBy | \BBx_1, \BBx_2) P(\BBx_1| \BBx_2)}{ P(\BBx_1|\BBy)} = P(\BBy| \BBx_2) \label{Eq_Ind1}
\end{eqnarray}

Since $P(\BBy=0|\BBx_2)+P(\BBy=1|\BBx_2) =  1$, Equation~\ref{Eq_Ind1} implies
\begin{eqnarray}
\frac{P(\BBy=0 | \BBx_1, \BBx_2) P(\BBx_1| \BBx_2)}{ P(\BBx_1|\BBy=0)} +
\frac{P(\BBy=1 | \BBx_1, \BBx_2) P(\BBx_1| \BBx_2)}{ P(\BBx_1|\BBy=1)} = 1 \nonumber \\
\Rightarrow \frac{P(\BBy=0 | \BBx_1, \BBx_2) P(\BBx_1| \BBx_2)}{ P(\BBx_1|\BBy=0)} + 
\frac{\left(1 - P(\BBy=0 | \BBx_1, \BBx_2)\right) P(\BBx_1| \BBx_2)}{ P(\BBx_1|\BBy=1)} = 1 \label{EQ_Posterior1}
\end{eqnarray}

Solving Equation~\ref{EQ_Posterior1} for $P(\BBy=0 | \BBx_1, \BBx_2)$, we obtain
\begin{eqnarray}
P(\BBy=0 | \BBx_1, \BBx_2) &=& \frac{P(\BBx_1|\BBy=0)}{P(\BBx_1|\BBx_2)} \cdot \frac{P(\BBx_1|\BBy=1)-P(\BBx_1|\BBx_2)}{P(\BBx_1|\BBy=1)-P(\BBx_1|\BBy=0)} \label{EQ_Surr1}
\end{eqnarray}

We have succeeded in writing $P(\BBy=0 | \BBx_1, \BBx_2)$ as a function of $P(\BBx_1|\BBx_2)$ and $P(\BBx_1|\BBy)$. This leads to a significant simplification of the learning task when a large amount of unlabeled data is available, especially if $\BBx_1$ is finite valued. The learning algorithm involves the following two steps.
\begin{itemize}
\item{Estimate the quantity $P(\BBx_1|\BBx_2)$ from only the unlabeled data, by building a predictor from the feature space $\mathcal{X}_2$ to the space $\mathcal{X}_1$. There is no restriction on the learning algorithm for this prediction task.}
\item{Estimate the quantity $P(\BBx_1|\BBy)$ from a smaller labeled sample by counting.} 
\end{itemize}
Thus, we can decouple the prediction problem into two separate tasks, one of which involves predicting $\BBx_1$ from the remaining features. In other words, $\BBx_1$ serves as a \emph{surrogate} for the class label. Furthermore, for the two steps above there is no necessity for complete samples. All the labeled examples can have the feature $\BBx_2$ missing.

\begin{figure}[htbp]
	\begin{tabular}{cc}
    \epsfxsize=2.5in\epsfysize=1.8in\epsfbox{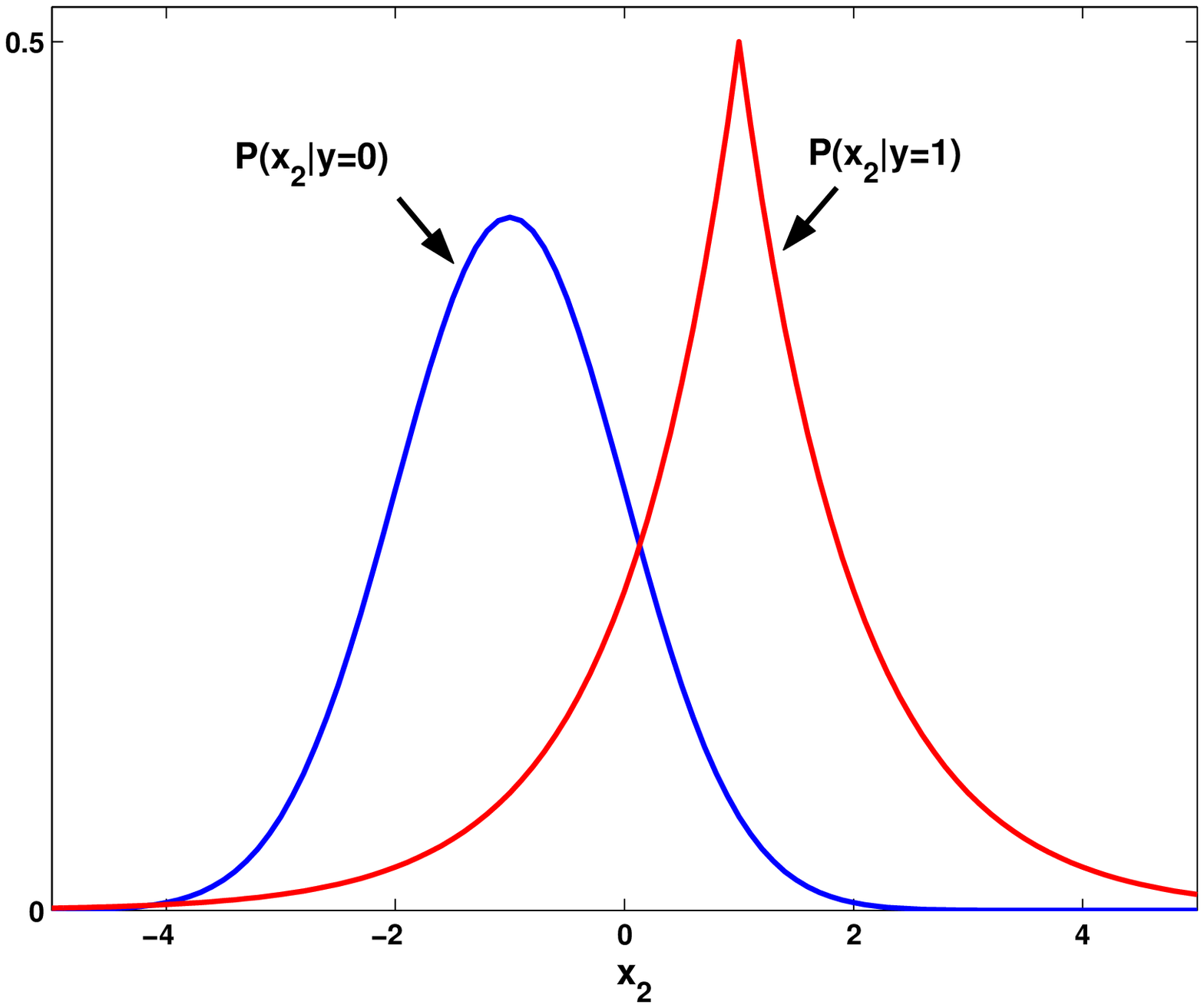} & \epsfxsize=2.5in\epsfysize=1.8in\epsfbox{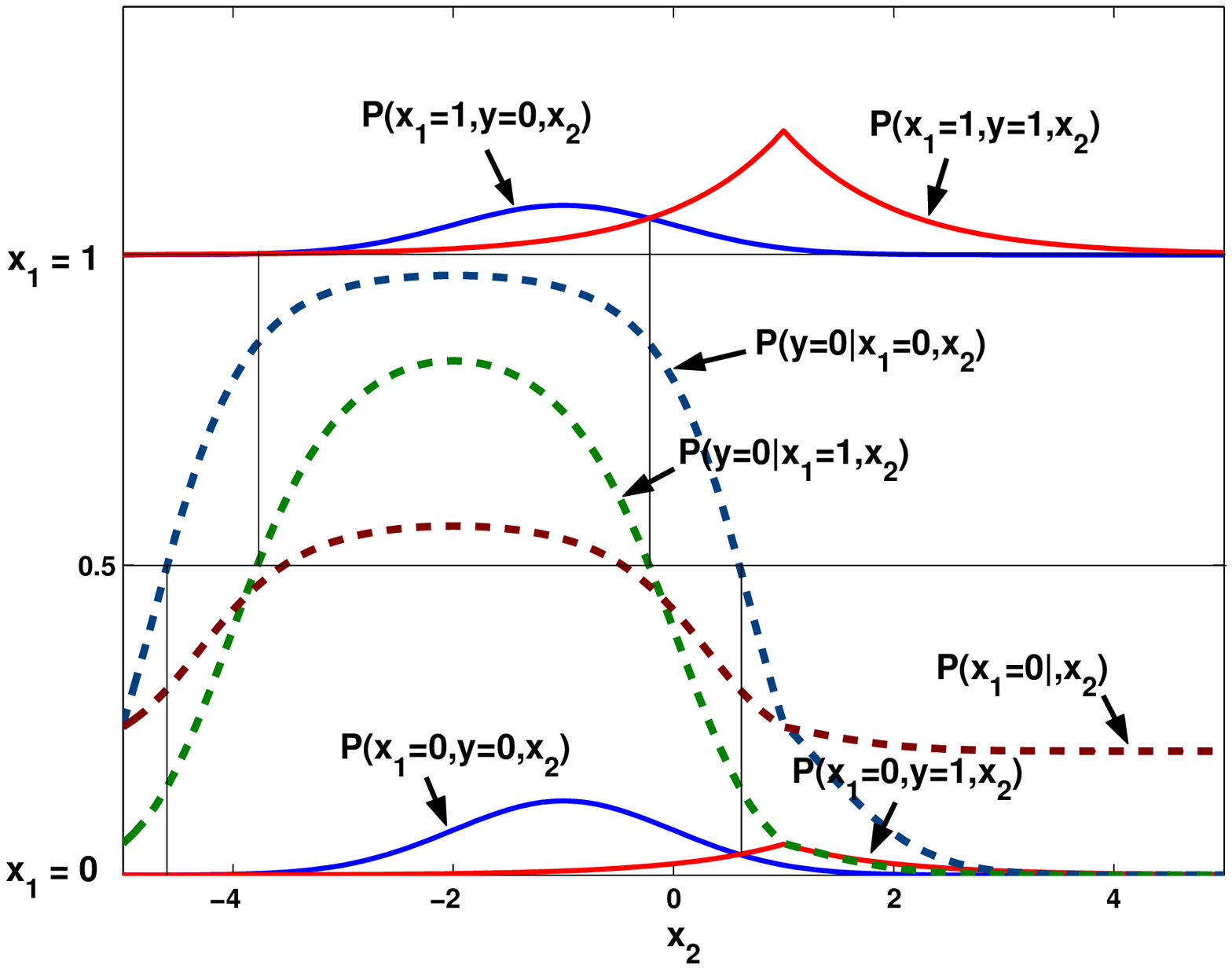}\\
    A&B
\end{tabular}
  \caption{A) Class-conditional probability distributions of the feature $\BBx_2$, B) the joint distributions and the posterior distributions of the class $\BBy$ and the surrogate class $\BBx_1$.}
  \label{fig_cond_x2_y}
\end{figure}

The following example illustrates the intuition behind surrogate learning.\\
----------------------------------------------------------------------\\
\emph{Example 1}\\
Consider a two-class problem, where $\BBx_1$ is a binary feature and $\BBx_2$ is a one dimensional real-valued feature. The class-conditional distribution of $\BBx_2$ for the class $\BBy=0$ is Gaussian, and for the class $\BBy=1$ is Laplacian as shown in Figure~\ref{fig_cond_x2_y}.A.

Because of the class-conditional feature independence assumption, the joint distribution $P(\BBx_1, \BBx_2, \BBy)$ can now be completely specified by fixing the joint probability $P(\BBx_1, \BBy)$. Let $P(\BBx_1 = 0, \BBy = 0) = 0.3$,  $P(\BBx_1 = 0, \BBy = 1) = 0.1$,  $P(\BBx_1 = 1, \BBy = 0) = 0.2$, and  $P(\BBx_1 = 1, \BBy = 1) = 0.4$. The full joint distribution is depicted in Figure~\ref{fig_cond_x2_y}.B. Also shown in Figure~\ref{fig_cond_x2_y}.B are the conditional distributions $P(\BBx_1=0|\BBx_2)$ and $P(\BBy=0|\BBx_1, \BBx_2)$. 

Assume that we have a classifier to decide between $\BBx_1 = 0$ and $\BBx_1 = 1$ from the feature $\BBx_2$. If this classifier is used to classify a sample that is from class $\BBy=0$, it will most likely be assigned the `label' $\hat{\BBx}_1 = 0$ (because, for class $\BBy=0$, $\BBx_1 = 0$ is more likely than $\BBx_1 = 1$), and a sample that is from class $\BBy=1$ is often assigned the `label' $\hat{\BBx}_1 = 1$. Consequently the classifier between $\BBx_1=0$ and $\BBx_1=1$ provides information about the true class label $\BBy$. This can also be seen in the similarities between the curves $P(\BBy=0|\BBx_1, \BBx_2)$ to the curve $P(\BBx_1|\BBx_2)$. 


\subsection{A Special Case}
\label{Sec_Special_Case}
We now specialize the above setting of the classification problem to the one realized in the applications we present later. We still wish to learn a classifier from $\mathcal{X} = \mathcal{X}_1 \times \mathcal{X}_2$ to the set of classes $\mathcal{Y} = \{0, 1\}$. We make the following assumptions.
\begin{enumerate}
\item{$\BBx_1$ is a binary random variable. That is, $\mathcal{X}_1 = \{0, 1\}$.}
\item{$P(\BBx_1, \BBx_2 | \BBy = 0)  = P(\BBx_1| \BBy =0)P(\BBx_2 | \BBy = 0)$. We require that the feature $\BBx_1$ be class-conditionally independent of the remaining features \emph{only} for the class $\BBy = 0$.}
\item{$P(\BBx_1 = 0, \BBy = 1) = 0$. This assumption says that $\BBx_1$ is a `100\% recall' feature for $\BBy=1$\footnote{This assumption can be seen to trivially enforce the independence of the features for class $\BBy=1$.}.} 
\end{enumerate}

Assumption~3 simplifies the learning task to the estimation of the probability $P(\BBy = 0|\BBx_1 = 1, \BBx_2)$ for every point $\BBx_2 \in \mathcal{X}_2$. We can proceed as before to obtain the expression in Equation~\ref{EQ_Surr1}.
\begin{eqnarray}
P(\BBy=0 | \BBx_1=1, \BBx_2) &=& \frac{P(\BBx_1 = 1|\BBy=0)}{P(\BBx_1 = 1 |\BBx_2)} \frac{P(\BBx_1 =1 |\BBy=1)-P(\BBx_1 =1 |\BBx_2)}{P(\BBx_1 =1 |\BBy=1)-P(\BBx_1 =1 |\BBy=0)}  \nonumber \\
&=&\frac{P(\BBx_1 = 1|\BBy=0)}{P(\BBx_1 = 1 |\BBx_2)}  \cdot \frac{1-P(\BBx_1 =1 |\BBx_2)}{1-P(\BBx_1 =1 |\BBy=0)} \nonumber \\
&=&\frac{P(\BBx_1 = 1|\BBy=0)}{P(\BBx_1 =0 |\BBy=0)}  \cdot \frac{P(\BBx_1 =0 |\BBx_2)}{(1-P(\BBx_1 = 0 |\BBx_2))} \label{EQ_Surr2} \end{eqnarray}

Equation~\ref{EQ_Surr2} shows that $P(\BBy=0 | \BBx_1=1, \BBx_2)$ is a monotonically increasing function of $P(\BBx_1 =0 |\BBx_2)$. This means that after we build a predictor from $\mathcal{X}_2$ to $\mathcal{X}_1$, we only need to establish the threshold on $P(\BBx_1 =0 |\BBx_2)$ to yield the optimum classification between $\BBy =0$ and $\BBy = 1$. Therefore the learning proceeds as follows.
\begin{itemize}
\item{Estimate the quantity $P(\BBx_1|\BBx_2)$ from only the unlabeled data, by building a predictor from the feature space $\mathcal{X}_2$ to the binary space $\mathcal{X}_1$. Again, there is no restriction on this prediction algorithm.}
\item{Use a small labeled sample to establish the threshold on $P(\BBx_1 =0 |\BBx_2)$.} 
\end{itemize}

In the unlabeled data, we call the samples that have $\BBx_1 = 1$ as the \emph{target} samples and those that have $\BBx_1 = 0$ as the \emph{background} samples. The reason for this terminology is clarified in Example 2.

\begin{figure}[htbp]
    \begin{minipage}[c]{0.58\linewidth}
      \epsfxsize=2.5in\epsfysize=1.8in\epsfbox{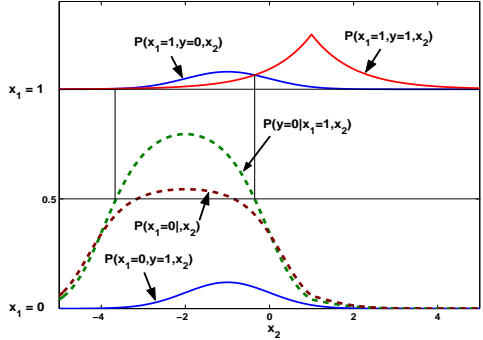}
    \end{minipage}\hfill
    \begin{minipage}[c]{0.38\linewidth}
  \caption{The joint distributions and the posterior distributions for the Example 2.}
  \label{fig_example2}
    \end{minipage}
\end{figure}

----------------------------------------------------------------------\\
\emph{Example 2}\\
We consider a problem with distributions $P(\BBx_2|\BBy)$ identical to Example 1 (Figure~\ref{fig_cond_x2_y}.A), except with the joint probability $P(\BBx_1, \BBy)$ given by $P(\BBx_1 = 0, \BBy = 0) = 0.3$,  $P(\BBx_1 = 0, \BBy = 1) = 0.0$,  $P(\BBx_1 = 1, \BBy = 0) = 0.2$, and  $P(\BBx_1 = 1, \BBy = 1) = 0.5$. The class-and-feature joint distribution is depicted in Figure~\ref{fig_example2}. Clearly, $\BBx_1$ is a 100\% recall feature for $\BBy =1$. 

Note that on the samples from the class $\BBy = 0$ it is impossible to determine whether it is a sample from the \emph{target} or \emph{background} better than random by looking at the $\BBx_2$ feature, whereas a sample from the positive class is always a \emph{target}. Therefore the \emph{background} samples serve to delineate the positive examples among the \emph{targets}.

\section{Related Work}
Although the idea of using unlabeled data to improve classifier accuracy has been around for several decades~\cite{NagyShelton66}, semi-supervised learning has received much attention recently due to impressive results in some domains. The compilation of chapters edited by Chappelle et al.~is an excellent introduction to the various approaches to semi-supervised learning, and the related practical and theoretical issues~\cite{ChaSchZie06}.

Identical to our setup, \emph{co-training} assumes that the features can be split into two class-conditionally independent sets or `views'~\cite{blum98combining}. Also assumed is the sufficiency of either view for accurate classification. The co-training algorithm iteratively uses the unlabeled data classified with high confidence by the classifier on one view, to generate labeled data for learning the classifier on the other. 

The intuition underlying co-training is that the errors caused by the classifier on one view are independent of the other view, hence can be conceived as uniform\footnote{Whether or not a label is erroneous is independent of the feature values of the latter view.} noise added to the training examples for the other view. Consequently, the number of label errors in a region in the feature space is proportional to the number of samples in the region. If the former classifier is reasonably accurate, the \emph{proportionally} distributed errors are `washed out' by the correctly labeled examples for the latter classifier. 

The main distinction of surrogate learning from co-training is the learning of a predictor from one view to the other, as opposed to learning predictors from both views to the class label. We can therefore eliminate the requirement that both views be sufficiently informative for reasonably accurate prediction. Furthermore, unlike co-training, surrogate learning has no iterative component. 

Ando and Zhang propose an algorithm to regularize the hypothesis space by simultaneously considering multiple classification tasks on the same feature space~\cite{DBLP:journals/jmlr/AndoZ05}. They then use their so-called \emph{structural learning} algorithm for semi-supervised learning of \emph{one} classification task, by the artificial construction of `related' problems on unlabeled data. This is done by creating problems of predicting \emph{observable} features of the data and learning the structural regularization parameters from these `auxiliary' problems and unlabeled data. More recently in~\cite{AndoZ07} they showed that, with conditionally independent feature sets predicting from one set to the other allows the construction of a feature representation that leads to an effective semi-supervised learning algorithm. Our approach directly operates on the original feature space and can be viewed another justification for the algorithm in~\cite{DBLP:journals/jmlr/AndoZ05}.

Castelli and Cover have studied the relative value of labeled and unlabeled samples for learning in a specialized setting where the class-conditional feature distributions are identifiable, and can be estimated from an unlabeled dataset~\cite{CastelliCover95,CastelliCover96}. After the mixture is identified from a large number of unlabeled samples and the classification boundary is defined, labeled examples are necessary only to specify the `orientation' of the boundary, i.e., to assign class labels to the regions in the feature space. We note the parallel in surrogate learning (cf. Equation~\ref{EQ_Surr2}), where a large amount of unlabeled data can be used to estimate the `terrain' $\frac{P(\BBx_1 =0 |\BBx_2)}{(1-P(\BBx_1 = 0 |\BBx_2))}$ and labeled data is necessary to choose the contour that defines the classification boundary.

Multiple Instance Learning (MIL) is a learning setting where training data is provided as positive and negative bags of samples~\cite{dietterich97solving}. A negative bag contains only negative examples whereas a positive bag contains at least one positive example. Surrogate learning can be viewed as artificially constructing a MIL problem, with the \emph{targets} acting as one positive bag and the \emph{backgrounds} acting as one negative bag (Section~\ref{Sec_Special_Case}). The class-conditional feature independence assumption for class $\BBy = 0$ translates to the identical and independent distribution of the negative samples in both bags.

\section{Two Applications}
We applied surrogate learning to problems in record linkage and natural language processing. We will explain below how  the learning problems in both the applications can be made to satisfy the assumptions in our second (100\% recall) setting.

\subsection{Record Linkage}
Record linkage is the process of identification and merging of records of the same entity in different databases or the unification of records in a single database, and constitutes an important component of data management. The reader is referred to~\cite{winkler95} for an overview of the record linkage problem, strategies and systems.

Our problem consisted of merging each of $\approx20000$ physician records, which we call the \emph{update database}, to the record of the same physician in a \emph{master database} of $\approx10^6$ records. The update database has fields that are absent in the master database and \emph{vice versa}. The fields in common include the \emph{name} (first, last and middle initial), several \emph{address} fields, \emph{phone}, \emph{specialty}, and the \emph{year-of-graduation}. Although the \emph{last name} and \emph{year-of-graduation} are consistent when present, the \emph{address}, \emph{specialty} and \emph{phone} fields have several inconsistencies owing to different ways of writing the address, new addresses, different terms for the same specialty, missing fields, etc. However, the \emph{name} and \emph{year} alone are insufficient for disambiguation. We had access to $\approx500$ manually matched update records for training and evaluation (about $40$ of these update records were labeled as unmatchable due to insufficient information).

The general approach to record linkage involves two steps: 1) \emph{blocking}, where a small set of candidate records is retrieved from the master record database, which contains the correct match with high probability, and 2) \emph{matching}, where the fields of the update records are compared to those of the candidates for scoring and selecting the match. We performed blocking by querying the master record database with the \emph{last name} from the update record. Matching was done by scoring a feature vector of similarities over the various fields. The feature values were either binary (verifying the equality of a particular field in the update and a master record) or continuous (some kind of normalized string edit distance between fields like \emph{street address}, \emph{first name} etc.).

The surrogate learning solution to our matching problem was set up as follows. We designated the binary feature of equality of \emph{year of graduation}\footnote{We believe that the equality of the middle intial would have worked just as well for $\BBx_1$.} as the surrogate label $\BBx_1$, and the remaining features are relegated to $\BBx_2$. The required conditions for surrogate learning are satified because 1) in our data it is highly unlikely for two records with different \emph{year- of-graduation} to belong to the same physician and 2) if it is known that the update record and a master record belong to two \emph{different} physicians, then knowing that they have the same (or different) \emph{year-of-graduation} provides no information about the other features. Therefore all the feature vectors with the binary feature indicating equality of \emph{year-of-graduation} are \emph{targets} and the remaining are \emph{backgrounds}.

First, we used feature vectors obtained from the records in all blocks from all 20000 update records to estimate the probability $P(\BBx_1|\BBx_2)$. We used logistic regression for this prediction task. For learning the logistic regression parameters, we discarded the feature vectors for which $\BBx_1$ was missing and performed mean imputation for the missing values of other features. Second, the probability $P(\BBx_1=1|\BBy = 0)$ (the probability that two different randomly chosen physicians have the same year of graduation) was estimated straightforwardly from the counts of the different years-of-graduation in the master record database. 

These estimates were used to assign the score $P(\BBy = 1|\BBx_1 = 1, \BBx_2)$ to the records in a block (cf. Equation~\ref{EQ_Surr2}). The score of $0$ is assigned to feature vectors which have $\BBx_1 = 0$. The only caveat is calculating the score for feature vectors that had missing $\BBx_1$. For such records we assign the score $P(\BBy=1|\BBx_2) = P(\BBy=1|\BBx_1 = 1, \BBx_2) P(\BBx_1 = 1|\BBx_2)$. We have estimates for both quantities on the right hand side. The highest scoring record in each block was flagged as a match if it exceeded some appropriate threshold.

We compared the results of the surrogate learning approach to a supervised logistic regression based matcher which used a portion of the manual matches for training and the remaining for testing. Table~1 shows the match precision and recall for both the surrogate learning and the supervised approaches. For the supervised algorithm, we show the results for the case where half the manually matched records were used for training and half for testing, as well as for the case where a fifth of the records of training and the remaining four-fifths for testing. In the latter case, every record participated in exactly one training fold but in four test folds.

\begin{table}[t]

\begin{center}
		\caption{Precision and Recall for record linkage. The surrogate learning algorithm had access to none of the manually matched records.}
	\begin{tabular}{ |l |r|r| r| }
    \hline
    &Training& Precision & Recall \\ 
    &proportion&  &  \\ \hline \hline
    Surrogate && 0.96 & 0.95 \\ \hline
    Supervised&0.5& 0.96 & 0.94 \\ \hline
    Supervised&0.2& 0.96 & 0.91 \\
    \hline
 \end{tabular}
\end{center}	\label{Table_Record_Linkage}
\end{table}

The results indicate that the surrogate learner performs better matching by exploiting the unlabeled data than the supervised learner with insufficient training data. The results although not dramatic are still promising, considering that the surrogate learning approach used \emph{none} of the training records. 

\subsection{Merger-Acquisition Sentence Classification}
Sentence classification is often a preprocessing step for event or relation extraction from text. One of the challenges posed by sentence classification is the diversity in the language for expressing the same event or relationship. We present a surrogate learning approach for constructing a sentence classifier that detects a \emph{merger-acquisition} (MA) event between two organizations in financial news (in other words, we find paraphrases for the MA event).

We assume that the unlabeled sentence corpus is time-stamped and named entity tagged with organizations. We further assume that a MA sentence must mention at least two organizations. Our approach to build the sentence classifier is the following. We first extract all the so-called \emph{source} sentences from the corpus that match a few high-precision seed patterns. An example of a seed pattern used for the MA event is `$<$ORG1$>$ acquired $<$ORG2$>$' (see Example 3 below).

We then extract every sentence in the corpus that contains at least two organizations, such that at least one of them matches an organization in the \emph{source} sentences, and has a time-stamp within a two month time window of the matching \emph{source} sentence. Of this set of sentences, all that contain \emph{two} or more organizations from the \emph{same} \emph{source} sentence are designated as \emph{target} sentences, and the rest are designated as \emph{background} sentences. 

We speculate that since an organization is unlikely to have a MA relationship with two different organizations in the same time period the \emph{backgrounds} are unlikely to contain MA sentences, and moreover the language of the non-MA \emph{target} sentences is indistinguishable from that of the \emph{background} sentences. 
To relate the approach to surrogate learning, we note that the binary ``organization-pair equality" feature (both organizations in the current sentence being the same as those in a \emph{source} sentence) serves as the `100\% recall' feature $\BBx_1$. The language in the sentence is the feature set $\BBx_2$. This setup satisfies the required conditions for surrogate learning because 1) if a sentence is about MA, the organization pair mentioned in it must be the same as that in a \emph{source} sentence, (i.e., if \emph{only} one of the organizations match those in a \emph{source} sentence, the sentence is unlikely to be about MA) and 2) if an unlabeled sentence is non-MA, then knowing whether or not it shares an organization with a \emph{source} does not provide any information about the language in the sentence. 

We then trained a support vector machine (SVM) classifier to discriminate between the \emph{targets} and \emph{backgrounds}. The feature set ($\BBx_2$) used for this task was a bag of word unigrams, bigrams and trigrams, generated from the sentences and selected by ranking the n-grams by the divergence of their distributions in the \emph{targets} and \emph{backgrounds}. The sentences were ranked according to the score assigned by the SVM (which is a proxy for $P(\BBx_1=1|\BBx_2)$). This score was then thresholded to obtain a classification between MA and non-MA sentences.

%


Example 3 below lists some sentences to illustrate the surrogate learning approach. 
Note that the \emph{targets} may contain both MA and non-MA sentences but the \emph{backgrounds}
are unlikely to be MA.

----------------------------------------------------------------------\\
\emph{Example 3}\\
{\small
{\bf Seed Pattern}\\
``offer for $<$ORG$>$"\\
{\bf Source Sentences}\\
1. $<$ORG$>$US Airways$<$ORG$>$ said Wednesday it will increase its {\bf offer for $<$ORG$>$Delta$<$ORG$>$}.\\
{\bf Target Sentences (SVM score)}\\
1.$<$ORG$>$US Airways$<$ORG$>$ were to combine with a standalone $<$ORG$>$Delta$<$ORG$>$. (1.0008563)\\
2.$<$ORG$>$US Airways$<$ORG$>$ argued that the nearly \$10 billion acquisition of $<$ORG$>$Delta$<$ORG$>$ would result in an efficiently run carrier that could offer low fares to fliers. (0.99958149)\\
3.$<$ORG$>$US Airways$<$ORG$>$ is asking $<$ORG$>$Delta$<$ORG$>$'s official creditors committee to support postponing that hearing. (-0.99914371)\\
{\bf Background Sentences (SVM score)}\\
1. The cities have made various overtures to $<$ORG$>$US Airways$<$ORG$>$, including a promise from $<$ORG$>$America West Airlines$<$ORG$>$ and the former $<$ORG$>$US Airways$<$ORG$>$. (0.99957752)\\
2. $<$ORG$>$US Airways$<$ORG$>$ shares rose 8 cents to close at \$53.35 on the $<$ORG$>$New York Stock Exchange$<$ORG$>$. (-0.99906444)
}

----------------------------------------------------------------------\\

We tested our algorithm on an unlabeled corpus of approximately 700000 financial news articles. We experimented with  five seed patterns ({\bf$<$ORG$>$ acquired $<$ORG$>$, $<$ORG$>$ bought $<$ORG$>$, offer for $<$ORG$>$, to buy $<$ORG$>$, merger with $<$ORG$>$}) which resulted in 870 \emph{source} sentences. The participants that were extracted from \emph{sources} resulted in approximately 12000 \emph{target} sentences and approximately 120000 \emph{background} sentences. For the purpose of evaluation, 500 randomly selected sentences from the \emph{targets} were manually checked leading to 330 being tagged as MA and the remaining 170 as non-MA. This corresponds to a 66\% precision of the \emph{targets}.

We then ranked the \emph{targets} according to the score assigned by the SVM trained to classify between the \emph{targets} and \emph{backgrounds}, and selected all the \emph{targets} above a threshold as paraphrases for MA. Table~3 presents the precision and recall on the 500 manually tagged sentences as the threshold varies. The results
indicate that our approach provides an effective way to rank the \emph{target} sentences according to their likelihood of being about MA.

\begin{table}[t]

\begin{center}
		\caption{Precision/Recall of surrogate learning on the MA sentence problem for various thresholds. The baseline of using all the \emph{targets} as paraphrases for MA has a precision of 66\% and a recall of 100\%.}
	\begin{tabular}{ |r|r|r|}
    \hline
    Threshold& Precision & Recall \\
    \hline 
    \hline
    0.0& 0.83 & 0.94 \\ \hline
    -0.2& 0.82 & 0.95 \\ \hline
    -0.8& 0.79 & 0.99  \\ \hline
 \end{tabular}
\end{center}	
\end{table}

We also evaluated the capability of the method to find paraphrases by conducting five separate experiments using each pattern in Table~2 individually as the only seed and counting the number of obtained sentences containing each of the other patterns (using a threshold of 0.0). We found that the method was effective in finding paraphrases that have very different language than the \emph{sources}. We do not provide the numbers due to space considerations.

Finally we used the paraphrase sentences, which were found by surrogate learning, to augment the training data for a MA
sentence classifier and evaluated its accuracy. We first built a SVM classifier only on a portion of the labeled \emph{targets} and used the remaining as the test set. This approach yielded an accuracy of 76\% on the test set (with two-fold cross validation). We then added all the \emph{targets} scored above a threshold by surrogate learning as positive examples (4000 positive sentences in all were added), and all the \emph{backgrounds} that scored below a low threshold as negative examples (27000 sentences), to the training data and repeated the two-fold cross validation. The classifier learned on the augmented training data improved the accuracy on the test data to 86\% . 

We believe that better designed features (than word n-grams) will provide paraphrases with higher precision and recall of the MA sentences found by surrogate learning.  To apply our approach to a new event extraction problem, the design step also involves the selection of the $\BBx_1$ feature such that the \emph{targets} and \emph{backgrounds} satisfy our assumptions.

\section{Conclusions}
We presented surrogate learning -- a simple semi-supervised learning algorithm that can be applied when the features  satisfy the required independence assumptions. We presented two applications, showed how the assumptions are satisfied, and presented empirical evidence for the efficacy of our algorithm. We expect that surrogate learning is sufficiently general to be applied in diverse domains, if the features are carefully designed. We are developing a version of the algorithm that allows the statistical independence assumption to be relaxed to mean independence.


\bibliographystyle{plain}
\bibliography{icml08}

\end{document}